# Deep Learning for Sleep Heart Rate Estimation from Accelerometers: Toward Population-Scale Cardiac Insight Without Optical Sensors

Tanbin Islam Rohan
*Department of Information Technology*
*Kennesaw State University*
Kennesaw, GA, USA
trohan1@students.kennesaw.edu

Pranjol Sen Gupta
*Department of Information Technology*
*Kennesaw State University*
Kennesaw, GA, USA
pgupta10@kennesaw.edu

Tanusree Debi
*School of Computing*
*University of Georgia*
Athens, GA, USA
td33362@uga.edu

Nazmus Sakib
*Department of Information Technology*
*Kennesaw State University*
Kennesaw, GA, USA
nsakib1@kennesaw.edu

***Abstract*—Large longitudinal cohorts often contain wrist accelerometry without optical heart-rate sensing, motivating recovery of cardiac information from motion signals already collected during sleep. We present SeqSmoother, a Transformer-based temporal corrector for sleep heart-rate (HR) estimation from wrist accelerometry. SeqSmoother combines spectral descriptors with an intermediate Nightbeat-derived frequency anchor and a physics-motivated sub-harmonic feature designed to identify harmonic frequency lock-on. All inference-time features are derived from wrist accelerometry, while ECG is used only to construct reference HR labels and training-label quality weights. We evaluate SeqSmoother using 13 participant-disjoint held-out folds and compare it with the official Nightbeat implementation under a matched 60-s window and 15-s step protocol. Across all out-of-fold predictions, SeqSmoother achieved a participant-macro MAE of 1.60 bpm. On Nightbeat-retained matched intervals, Nightbeat achieved lower absolute error than SeqSmoother (0.615 versus 1.091 bpm), while SeqSmoother provided estimates over a larger portion of the eligible recording; Nightbeat produced final estimates for 72.85% of the SeqSmoother-eligible out-of-fold grid. Separately, the proposed sub-harmonic ratio achieved an AUROC of 0.972 for identifying reference-defined harmonic lock-on candidates. These findings reveal an accuracy–availability trade-off between learned temporal modeling and quality-gated signal processing while providing empirical support for a physics-informed approach to identifying frequency-tracking failures in accelerometer-based sleep HR estimation.***

***Index Terms*—heart rate estimation, ballistocardiogram, accelerometer, deep learning, sleep monitoring, wearable sensors, harmonic lock-on.**

## I. Introduction

Wrist-worn accelerometers are widely used in large longitudinal health studies to record physical activity and sleep. The U.K. Biobank [1], for example, collected week-long wrist accelerometry from roughly 100,000 participants. Such cohorts often contain rich longitudinal health records but no continuous optical heart-rate (HR) channel. Recovering HR from accelerometry already available in these cohorts could therefore provide an additional physiological signal for future retrospective analyses, provided that the reconstruction method is first validated against a suitable cardiac reference.

The mechanical recoil of each heartbeat produces the ballistocardiogram (BCG), a vibration measurable by accelerometers [2], [3]. At the wrist, this signal can be overwhelmed by voluntary motion, whereas sleep provides longer periods with relatively little movement [4], [5]. Earlier wrist approaches either relied on movement-free periods or required stronger external reference information [5], [6]. More recently, Nightbeat [7] combined frequency-domain tracking, motion-artifact rejection, and fixed temporal post-processing to obtain strong HR estimation on uncontrolled at-home sleep recordings.

Nightbeat therefore provides a demanding signal-processing (SP) baseline. Rather than asking whether a learned model should replace such a pipeline, we investigate whether learned temporal modeling can provide complementary behavior. We present *SeqSmoother*, a Transformer-based temporal corrector that uses a Nightbeat-derived frequency estimate as an anchor together with spectral and temporal information from wrist accelerometry. SeqSmoother also incorporates a physics-motivated *sub-harmonic ratio* designed to identify frequency-tracker lock-on near twice the true cardiac fundamental.

The evaluation treats the participant as the independent statistical unit. SeqSmoother is evaluated on 13 participant-disjoint held-out folds, and the official Nightbeat implementation is re-evaluated under a matched 60-s/15-s temporal protocol. The results show that Nightbeat achieves lower absolute error on intervals retained by its quality-control pipeline, whereas SeqSmoother provides predictions over a broader portion of the eligible recording. Separately, we evaluate the proposed sub-harmonic ratio against reference-defined harmonic lock-on events. We therefore interpret the findings as an *accuracy–availability trade-off* between learned temporal modeling and quality-gated signal processing, together with empirical support for a physics-informed diagnostic of frequency-tracking failures.

**Contributions.** This paper makes three contributions:

1) We develop *SeqSmoother*, a Transformer-based temporal corrector for accelerometer-only sleep HR estimation that combines a Nightbeat-derived frequency anchor with spectral and temporal features from wrist accelerometry.
2) We introduce a physics-motivated sub-harmonic ratio $\eta_{1/2}$ for identifying harmonic frequency lock-on and empirically validate it using a pooled reference-based diagnostic, achieving an AUROC of 0.972 across 337 candidate lock-on windows from 7 held-out participants.
3) We provide a participant-level matched comparison between SeqSmoother and the official Nightbeat implementation, including participant bootstrap uncertainty, paired testing, within-participant-centered correlation, and explicit alignment, retention, and prediction-availability analysis.

## II. Related Work

### A. HR Estimation from Wearables and the Wrist BCG

Most consumer wearables estimate HR optically via photoplethysmography (PPG), which can degrade under motion artifacts [8]. Accelerometer-based cardiac sensing avoids the optical front end by measuring mechanical cardiac motion. The characteristic BCG waveform is more readily observed on the chest [2], in instrumented furniture, or under a mattress [3]. At the wrist, cardiac vibrations are much weaker relative to movement, so sleep has been an important operating regime [4], [5]. BioInsights [6] extracts HR from relatively still wearable motion segments; PWR [5] reconstructs pulse-related information using filtering and Hilbert-domain processing; and the jerk-based method of Weaver et al. [4] differentiates wrist acceleration before spectral analysis. Nightbeat [7] combines frequency tracking, artifact rejection, and temporal post-processing and serves as the primary signal-processing reference in this study.

### B. Learned Models for Physiological Signals

Convolutional, recurrent, and attention-based models are widely used for physiological-signal estimation. Long short-term memory networks [9] and residual convolutional architectures [10] have been effective for sequential and raw biosignals, while Transformers [11] provide a flexible mechanism for modeling longer temporal dependencies. Recent work has applied Transformer architectures to BCG-based HR estimation [12] and to PPG/rPPG HR estimation [13], [14]. However, these studies differ in sensor modality, acquisition setting, and evaluation protocol from the wrist-accelerometer-only sleep setting considered here, so their reported performance is not directly comparable. Prior work also shows that carefully designed signal-processing methods can remain highly competitive with learned approaches [15], while uncertainty-aware modeling can improve robustness in wearable HR estimation [8].

In this work, the Transformer serves as the temporal modeling backbone rather than as an architectural novelty. The contribution instead lies in combining learned temporal modeling with an accelerometer-derived frequency anchor and physics-motivated spectral information for sleep HR estimation.

### C. Datasets for Wrist-BCG HR During Sleep

Several sleep datasets support accelerometer-based HR analysis, but many are recorded in controlled or laboratory environments [5]. NightbeatDB [7] instead contains at-home recordings with synchronized wrist accelerometry and ECG reference data. We use it because it provides the reference signal needed for method validation while preserving realistic home-sleep variability.

## III. The NightbeatDB Dataset

We use NightbeatDB [7], a dataset of full-night, at-home recordings from 42 participants. Each participant wore a wrist-mounted triaxial accelerometer (Axivity AX3, $f_a$ = 100 Hz) on the dominant wrist and an ECG chest belt (movisens Ecg-Move 4, $f_e$ = 1,024 Hz) providing the cardiac reference. Participants followed normal sleep routine and self-reported bed and wake times and bed-sharing status. After synchronization, 38 of the 42 recordings yielded usable aligned accelerometer–ECG data. The finalized windowing, Welch-feature extraction, and intermediate Nightbeat-anchor pipeline was successfully executed across all 38 usable recordings to verify cohort-wide pipeline applicability and data availability. This all-recording pass is a preprocessing consistency check only; formal model-performance inference is based on the 13 participant-disjoint held-out folds reported below. All model splits are performed at the participant level so that windows from a held-out participant never appear in training or validation.

## IV. Method

We investigate *SeqSmoother*, a Transformer-based temporal corrector (Fig. 1) that operates on a compact spectral feature representation and learns to refine an intermediate Nightbeat-derived frequency anchor using temporal context. SeqSmoother is evaluated under a participant-disjoint held-out protocol, with the statistical analysis based on a matched comparison against the official Nightbeat implementation.

### A. Data Preparation and Windowing

Let $\mathbf{x}(t) = [x(t), y(t), z(t)]^\top$ denote the triaxial accelerometer signal and let $B = \{b_1, \ldots, b_M\}$ be the ECG beat timestamps. Each full-night recording is partitioned into overlapping 180-s chunks with a 120-s overlap (60-s chunk stride). The accelerometer, ECG validity information, and beat timestamps are stored in aligned per-participant files.

For SeqSmoother, 60-s windows are extracted at the same 15-s step (75% overlap). Duplicate windows created by chunk overlap are removed using their absolute timestamps before sequence construction.

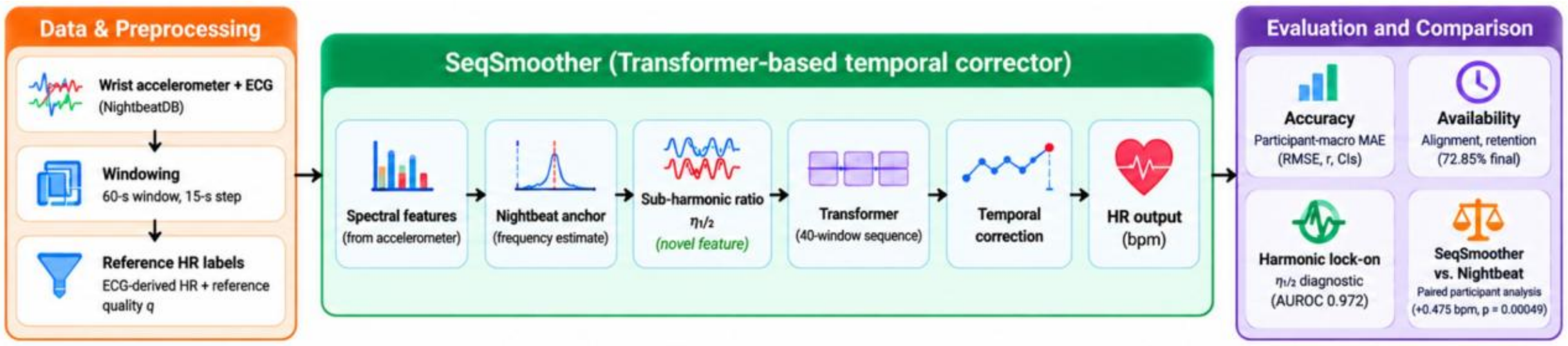

Fig. 1. Study framework for SeqSmoother, the Transformer-based temporal corrector used for accelerometer-only sleep HR estimation. Wrist accelerometry is segmented into overlapping 60-s windows and transformed into spectral features, including an intermediate Nightbeat-derived frequency anchor and the proposed sub-harmonic ratio $\eta_{1/2}$. A 40-window Transformer sequence models temporal context and produces the final HR estimate. ECG provides the reference HR labels and training-label quality information but is not used at inference. Evaluation includes participant-level accuracy and prediction availability, the sub-harmonic lock-on diagnostic, and a matched participant-level comparison with the official Nightbeat implementation.

### *B. HR Label Construction*

For a window spanning $[t_s, t_e)$, the HR label is derived from ECG beat timestamps within the same interval. Inter-beat intervals are

$$\mathrm{IBI}_i = \frac{b_{i+1} - b_i}{f_e}. \tag{1}$$

Intervals outside $0.3 < \mathrm{IBI}_i < 2.0$ s are excluded. If $N'$ valid intervals remain, the reference HR is

$$y = \frac{60}{\frac{1}{N'}\sum_{i=1}^{N'} \mathrm{IBI}_i}. \tag{2}$$

Undefined labels or values outside $[30, 120]$ bpm are discarded.

### *C. ECG Reference-Quality Score*

Each window receives an ECG reference-quality score $q \in [0, 1]$, defined as the fraction of ECG samples marked valid in that interval,

$$q = \frac{1}{e - s}\sum_{t=s}^{e-1} \mathbf{1}(v_t > 0). \tag{3}$$

Importantly, $q$ describes the reliability of the *ECG reference label*; it is not an accelerometer-motion or accelerometer-signal-quality measure. No hard $q$ threshold is used for training, validation, or testing ($q_{\min} = 0$). Instead, $q$ only down-weights less reliable training labels. We do not use $q$ as a proxy for wrist motion or accelerometer quality, and it is not an inference-time model input.

### *D. SeqSmoother*

SeqSmoother uses a 26-dimensional feature vector for each 60-s window and a Transformer [11] to model temporal context. The features combine an intermediate Nightbeat-derived frequency anchor, spectral descriptors, the proposed sub-harmonic descriptor, and temporal change information.

*1) Nightbeat-derived anchor:* For each window, the feature pipeline produces an intermediate Nightbeat-derived HR anchor $\hat{h}_{\mathrm{NH}}$ and an anchor-availability/quality indicator. If the intermediate anchor is unavailable, a spectral fallback estimate is used and the feature representation records that fallback state. This intermediate anchor is a SeqSmoother input and is distinct from the *final* official Nightbeat estimate used in the matched baseline evaluation. At inference, the anchor and all other SeqSmoother features are computed from wrist accelerometry; ECG-derived $q$ is used only as a training-label weight.

We denote the accelerometer-derived Nightbeat anchor availability/quality indicator by $q_{\mathrm{NH}}$. This quantity is distinct from the ECG reference-quality score $q$ defined in Section IV-C; $q_{\mathrm{NH}}$ is available at inference, whereas ECG-derived $q$ is used only as a training-label weight.

*2) Spectral features:* For each 60-s window, the vector magnitude $m(t) = \|\mathbf{x}(t)\|_2$ is bandpass filtered to 0.5–4.0 Hz. The power spectral density is then computed over the *full 60-s window* using Welch's method with a Hann window, $n_{\mathrm{perseg}} = 512$ samples, $n_{\mathrm{overlap}} = 256$, and $n_{\mathrm{FFT}} = 512$. From this spectrum we derive an 18-bin log-normalized PSD representation over the BCG band, a spectral-peak HR estimate, a band signal-to-noise descriptor, and the second-harmonic ratio.

*3) Sub-harmonic lock-on feature:* A common frequency-tracking failure occurs when the estimated frequency is near twice the true cardiac fundamental. The conventional second-harmonic ratio does not directly test for energy at one-half of a potentially locked anchor. During development, we explored several sub-harmonic formulations, including alternative half-frequency spectral ratios and an explicit half-HR candidate. We retained the ratio below because it directly compares spectral evidence at the hypothesized fundamental with the power at the current anchor, giving a simple normalized measure that is closely tied to the expected frequency-doubling failure mode.

We therefore define

$$\eta_{1/2} = \frac{P(\hat{h}_{\mathrm{NH}}/120)}{P(\hat{h}_{\mathrm{NH}}/60) + \epsilon}, \qquad \epsilon = 10^{-10}. \tag{4}$$

If the anchor is near twice the true HR, the numerator samples spectral energy near the true cardiac fundamental and $\eta_{1/2}$ is expected to increase. Fig. 2(a) illustrates this mechanism: when $\hat{h}_{\mathrm{NH}} \approx 2y$, the half-anchor frequency $\hat{h}_{\mathrm{NH}}/120$ falls near the true cardiac fundamental, whereas $\hat{h}_{\mathrm{NH}}/60$ corresponds to the doubled anchor. Elevated relative power at the half-anchor frequency therefore provides a physically interpretable cue for harmonic lock-on. A companion half-HR candidate $h_{1/2} = \hat{h}_{NH}/2$ provides an explicit alternative estimate.

*4) Temporal delta feature:* A clipped first difference

$$\Delta_k = \text{clip}\left(\frac{\hat{h}_{NH,k} - \hat{h}_{NH,k-1}}{15}, -2, 2\right) \quad (5)$$

encodes rapid anchor changes over the 15-s step.

*5) Full feature vector:* The 26-dimensional feature vector is

$$\mathbf{f} = [\tilde{h}_{NH}, q_{NH}, \tilde{h}_{spec}, \widetilde{SNR}, \tilde{\eta}_2, \tilde{\eta}_{1/2}, \tilde{h}_{1/2}, \mathbf{p}^\top, \Delta]^\top, \quad (6)$$

where $\mathbf{p} \in \mathbb{R}^{18}$ is the Welch PSD representation and tildes denote feature normalization. Training-only feature augmentation applies anchor jitter with $\sigma = 2.5$ bpm ($p = 0.60$), spectral-HR jitter with $\sigma = 1.5$ bpm ($p = 0.50$), sub-harmonic jitter with $\sigma = 0.05$ ($p = 0.30$), Gaussian PSD noise with $\sigma = 0.05$ ($p = 0.40$), and random masking of 1–5 adjacent PSD bins ($p = 0.30$).

*6) Transformer architecture:* Sequences contain $K = 40$ consecutive windows at a 15-s step. The token centers span 585 s, while the underlying 60-s window support extends the raw-data coverage to approximately 645 s. Sequences with a gap larger than 45 s between adjacent token timestamps are discarded.

Each token is projected to 64 dimensions, normalized, and combined with a learned positional embedding. Two pre-layer-normalized Transformer encoder layers use four attention heads, feed-forward dimension 256, and dropout 0.15. The head produces a bounded HR prediction for every token. All positions are supervised during training; the final token is used for the primary out-of-fold test prediction. The present study evaluates retrospective sleep reconstruction and does not claim prospective streaming performance. The final SeqSmoother contains 106,625 trainable parameters.

### E. Loss Function

For prediction $\hat{y}$ and reference $y$, let $a = |\hat{y} - y|$. The final SeqSmoother objective combines a Huber term, an additional tail penalty for errors above 3 bpm, ECG-reference-quality weighting, and a Pearson-correlation term:

$$w = 0.30 + 0.70q, \quad (7)$$

$$\mathcal{L}_{reg} = \mathbb{E}\left[w\left(\ell_{1.5}(\hat{y} - y) + 0.08\max(a - 3, 0)^2\right)\right], \quad (8)$$

$$\mathcal{L}_{SS} = \mathcal{L}_{reg} + 0.15(1 - r_{batch}). \quad (9)$$

The ECG-derived $q$ is used only during training and is never required at deployment.

### F. Training Configuration

SeqSmoother uses AdamW with learning rate $3 \times 10^{-4}$, weight decay $10^{-3}$, batch size 128, five warm-up epochs, and cosine decay to $10^{-6}$. Training is capped at 120 epochs, with a minimum of 15 epochs and early stopping patience of 30 epochs on validation MAE. Gradient norm is clipped to 1.0. Stochastic Weight Averaging is *not* used in the analysis; each fold is evaluated using its ordinary best-validation checkpoint.

### G. Participant-Level Evaluation Protocol

The formal analysis comprises the first 13 completed test folds in a pre-generated frozen participant order. The participant order, preprocessing pipeline, model configuration, loss function, and training settings were fixed before evidence accumulation and were unchanged across these folds; no test participant was added, removed, replaced, or selected based on held-out performance. In each fold, the other 37 usable participants are deterministically divided into 29 training and 8 validation participants; the test participant is never used for feature normalization, model selection, or early stopping. Separately, the finalized preprocessing and feature-generation pipeline was run across all 38 usable recordings as a descriptive applicability check. These additional recordings are not counted as held-out model-performance folds and do not increase the inferential sample size.

The official ETH-SIPLAB Nightbeat implementation is used for the baseline comparison. Its default/published-style configuration is used only as an implementation sanity check. For the direct comparison, the official implementation is evaluated using 60-s windows with 75% overlap (15-s step), matching SeqSmoother's temporal grid. Final Nightbeat estimates are aligned to SeqSmoother endpoints within 0.11 s. Direct accuracy comparisons use only aligned timestamps at which the official Nightbeat final pipeline retains a prediction. Both methods are scored against the same canonical ECG-reference HR vector on those matched intervals.

We report SeqSmoother performance on all eligible out-of-fold windows, a strict paired comparison on Nightbeat-retained matched intervals, and SeqSmoother performance on aligned intervals rejected by Nightbeat. Prediction availability is measured relative to the 20,792-endpoint SeqSmoother-eligible out-of-fold grid; timestamp alignment and Nightbeat retention conditional on alignment are reported separately because the final Nightbeat pipeline may abstain.

### H. Statistical Analysis

The participant is the independent unit for population inference. We report participant-macro MAE, participant-specific Pearson correlation summarized by median and IQR, and participant-bootstrap 95% confidence intervals using 5,000 participant resamples. For the matched comparison, paired participant-level differences are bootstrapped and tested with the Wilcoxon signed-rank test. Pooled window-level metrics are reported only as descriptive summaries.

To evaluate temporal association without inflation from between-participant HR differences, we also compute a within-participant-centered Pearson correlation: the participant-specific mean is removed from both prediction and reference before concatenating the centered values.

### I. Reference-Based Harmonic Lock-On Diagnostic

The proposed sub-harmonic feature is evaluated directly against the specific mechanism it was designed to represent rather than inferred only from aggregate regression performance. Among windows with a genuine intermediate

Nightbeat-derived anchor, a positive candidate harmonic lock-on window is defined as

$$|\hat{h}_{\mathrm{NH}} - 2y| \leq 5 \text{ bpm}, \tag{10}$$

where $y$ is the ECG-reference HR; all other eligible anchor windows form the negative class. The ability of $\eta_{1/2}$ to discriminate the two classes is quantified by AUROC. Because the 60-s windows overlap at a 15-s step and positive candidate windows occur in only seven participants, this AUROC is reported as a *pooled mechanistic diagnostic*, not as participant-level inference or as 337 independent observations. The analysis is not a clinical arrhythmia detector, and we do not translate its diagnostic AUROC into an isolated overall-MAE gain for the complete multifeature model.

## V. Results

### A. Primary Participant-Level SeqSmoother Evaluation

Across the 13 held-out participants, SeqSmoother produced 20,792 out-of-fold predictions. The participant-macro MAE was $1.601 \pm 0.527$ bpm, with a participant-bootstrap 95% CI of $[1.353, 1.899]$ bpm. The pooled descriptive MAE and RMSE were 1.613 and 3.154 bpm, respectively. The median participant-specific Pearson correlation was $r = 0.771$ [IQR: 0.683–0.833], while the pooled within-participant-centered correlation across all out-of-fold predictions was $r = 0.765$. These windows quantify evaluated signal coverage; they are not treated as independent inferential samples. The same finalized preprocessing and anchor pipeline was successfully executed for all 38 usable recordings as a descriptive applicability check, but only the 13 participant-disjoint held-out folds are used for model-performance inference.

### B. Matched SeqSmoother–Nightbeat Comparison

Of the 20,792 SeqSmoother out-of-fold endpoints, 19,478 (93.68%) aligned to an official Nightbeat candidate timestamp. Nightbeat retained a final estimate for 15,146 of those aligned intervals, corresponding to 77.76% retention among aligned candidates and 72.85% final prediction availability relative to the complete SeqSmoother out-of-fold grid.

On the exact 15,146 Nightbeat-retained matched intervals, participant-macro MAE was 1.091 bpm for SeqSmoother and 0.615 bpm for Nightbeat. The paired participant-level difference was

$$\Delta\mathrm{MAE} = \mathrm{MAE}_{\mathrm{Seq}} - \mathrm{MAE}_{\mathrm{NB}} = +0.475 \text{ bpm}, \tag{11}$$

with a 95% participant-bootstrap CI of $[+0.315, +0.638]$ bpm and Wilcoxon $p = 0.00049$, favoring Nightbeat in absolute error. Matched within-participant-centered correlations were $r = 0.878$ for SeqSmoother and $r = 0.854$ for Nightbeat; this difference is reported descriptively and is not used to claim statistically established temporal superiority.

The published Nightbeat value of 0.88 bpm [7] is retained only as external historical context; it is not used for paired inference because it was generated under a different protocol.

TABLE I
Participant-level evaluation on 13 independent held-out participants. Matched MAE and centered $r$ use the same 15,146 Nightbeat-retained intervals; final prediction availability is measured relative to the 20,792-endpoint SeqSmoother-eligible out-of-fold grid.

| Method / setting | MAE (bpm) | Centered $r$ | Availability |
|---|---|---|---|
| SeqSmoother, all OOF | 1.601* | 0.765 | 100.0% |
| SeqSmoother, matched | 1.091* | 0.878 | – |
| Nightbeat, matched | 0.615* | 0.854 | 72.85% |

*Participant-macro MAE. Nightbeat timestamp alignment is 93.68%; retention among aligned candidates is 77.76%, corresponding to 72.85% final availability on the SeqSmoother-eligible OOF grid.

### C. Accuracy–Availability and Failure Analysis

The 4,332 aligned candidate intervals rejected by the final Nightbeat pipeline were more difficult for SeqSmoother than the retained subset. On these rejected intervals, SeqSmoother obtained MAE 2.656 bpm, RMSE 4.437 bpm, and pooled descriptive $r = 0.815$. This quantifies the expected consequence of Nightbeat's quality gating rather than implying that wider availability is automatically preferable. Across the 13 participants, SeqSmoother all-window MAE ranged from 1.034 to 2.744 bpm, while Nightbeat conditional retention ranged from 64.9% to 89.0%. Nightbeat rejection is an operational difficult-window indicator, not a pure wrist-motion label, and ECG-derived $q$ is treated strictly as reference-label quality.

Reference construction was also checked directly between the SeqSmoother and official Nightbeat evaluation paths. Across 19,478 aligned intervals, 99.95% of paired ECG-reference values agreed within 0.01 bpm, with median and 95th percentile absolute differences equal to 0.000 bpm. Thus, the matched accuracy difference is not explained by inconsistent ECG target construction. NightbeatDB is dominated by typical nocturnal HR values, so we do not form unstable extreme-HR strata or generalize the results to arrhythmia or markedly elevated nocturnal HR.

### D. Sub-Harmonic Lock-On Diagnostic

Across the 13 held-out participants, the pooled reference-based diagnostic identified 337 candidate harmonic lock-on windows in 7 participants. As shown in Fig. 2(b), diagnostic performance remained high as evidence accumulated across the completed participant-disjoint folds, with the final pooled AUROC reaching 0.972 for $\eta_{1/2}$. The accumulation pattern also shows that candidate lock-on events were concentrated in a subset of participants rather than uniformly distributed across the cohort. Because the 60-s windows overlap at a 15-s step and positive candidates occur in only seven participants, this AUROC is interpreted as a pooled mechanistic diagnostic rather than participant-level inference.

Together, Fig. 2(a) and Fig. 2(b) connect the physical motivation of the feature with its observed diagnostic behavior. The result supports the hypothesis that spectral energy near one-half of a potentially doubled anchor is informative for identifying harmonic lock-on. However, it does not establish that $\eta_{1/2}$ alone produces a specific reduction in overall SeqSmoother MAE, because the complete model jointly uses spectral, anchor, and temporal features.

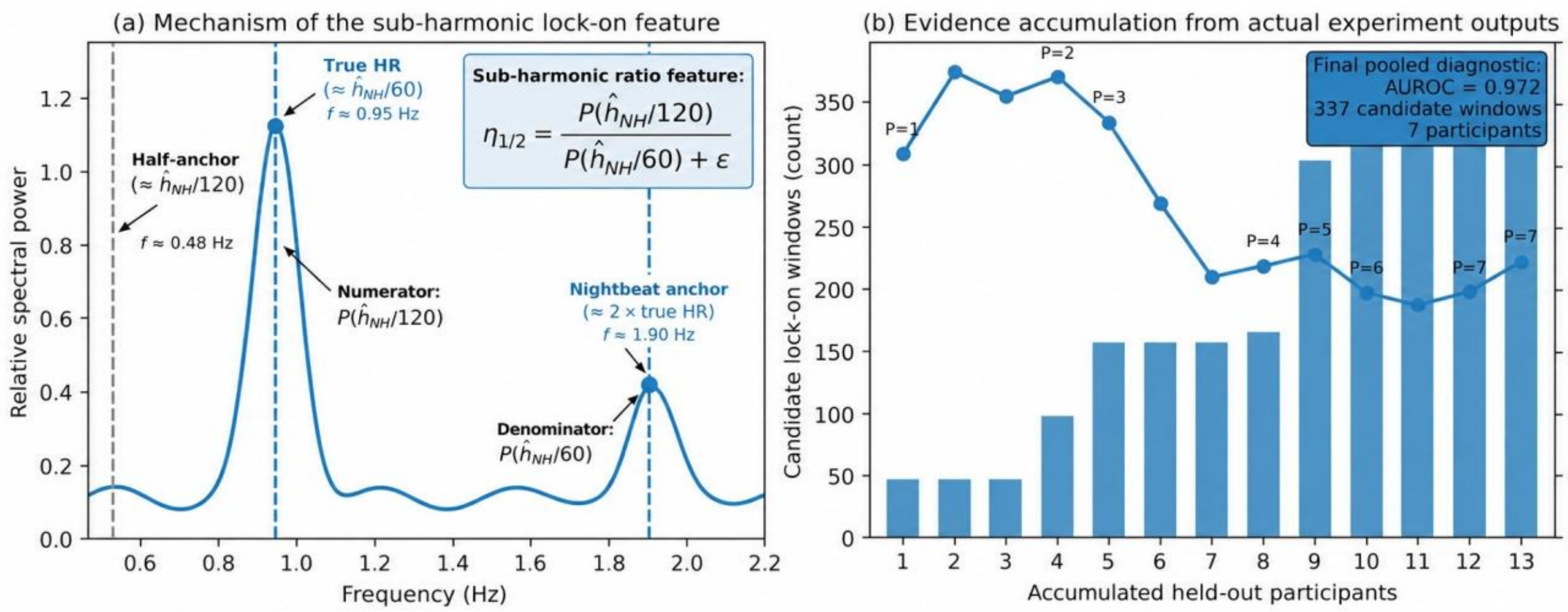

Fig. 2. Sub-harmonic lock-on mechanism and pooled diagnostic evidence. **(a)** Schematic illustration of the proposed sub-harmonic ratio $\eta_{1/2}$. When the intermediate Nightbeat-derived anchor $\hat{h}_{\mathrm{NH}}$ locks near twice the true cardiac fundamental, spectral power near $\hat{h}_{\mathrm{NH}}/2$ corresponds to the true HR frequency, causing $\eta_{1/2}$ to increase. The spectrum in panel (a) is illustrative rather than a measured participant example. **(b)** Evidence accumulation across the 13 completed participant-disjoint test folds. Bars show the cumulative number of reference-defined candidate lock-on windows, while the line shows the pooled AUROC of $\eta_{1/2}$. The final diagnostic includes 337 candidate windows from 7 participants and achieves AUROC 0.972. Because the 60-s windows overlap at a 15-s step, this result is interpreted as a pooled mechanistic diagnostic rather than participant-level inference.

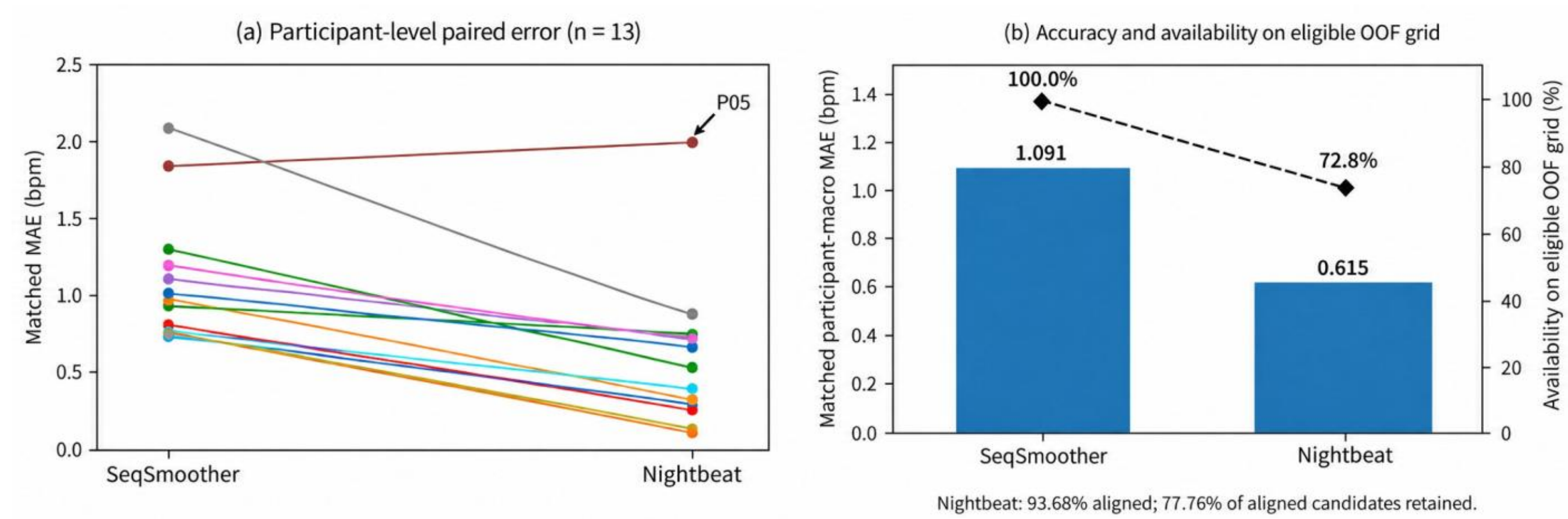

Fig. 3. Participant-level matched error and prediction availability. **(a)** Paired matched MAE for the 13 independent held-out participants; Nightbeat has lower matched MAE in 12 of 13 participants. **(b)** Matched participant-macro MAE is shown together with final prediction availability on the 20,792-endpoint SeqSmoother-eligible out-of-fold grid: 100% for SeqSmoother and 72.85% for Nightbeat. Nightbeat aligns to 93.68% of the grid and retains 77.76% of those aligned candidates after quality gating.

## VI. Discussion

The participant-level analysis changes the main interpretation of the study. Official Nightbeat is more accurate in absolute terms on intervals retained by its quality-control pipeline: matched participant-macro MAE is 0.615 bpm for Nightbeat versus 1.091 bpm for SeqSmoother, with a paired difference of +0.475 bpm favoring Nightbeat. SeqSmoother should therefore not be interpreted as a replacement for the engineered signal-processing pipeline; it is a hybrid learned temporal corrector built around an accelerometer-derived intermediate Nightbeat anchor.

The comparison instead exposes an accuracy–availability trade-off. Final Nightbeat predictions are available for 72.85% of the SeqSmoother-eligible out-of-fold grid; conditional on timestamp alignment, 77.76% of Nightbeat candidates are retained. SeqSmoother continues to provide predictions in regions where final Nightbeat abstains. These rejected regions are substantially more difficult—SeqSmoother's MAE there rises to 2.656 bpm—but the pooled descriptive correlation of $r = 0.815$ suggests that these intervals may still contain temporal HR information. This value is not used for participant-level inference. On the strict matched subset, SeqSmoother has a slightly higher within-participant-centered correlation (0.878 versus 0.854), which is reported descriptively rather than as statistically established superiority.

Fig. 3 summarizes the participant-level matched comparison and the corresponding accuracy–availability trade-off. As shown in Fig. 3(a), Nightbeat achieves lower matched MAE for 12 of the 13 held-out participants, while Fig. 3(b) shows that this lower error is accompanied by lower final prediction availability (72.85% versus 100% on the SeqSmoother-eligible OOF grid).

The sub-harmonic analysis provides a separate mechanistic contribution. A pooled diagnostic AUROC of 0.972 across 337 reference-defined candidate lock-on windows from seven participants supports the hypothesis that energy at one-half of a potentially doubled anchor is informative for harmonic frequency lock-on. Because windows overlap and the full SeqSmoother jointly uses spectral, anchor, and temporal information, we do not attribute a specific portion of total regression improvement uniquely to $\eta_{1/2}$.

These findings motivate future hybrid systems that preserve high-confidence Nightbeat estimates while using learned temporal modeling for uncertain or rejected regions. Such systems require prospective and external validation before clinical or population-level claims can be made.

## VII. Limitations and Future Work

First, formal model-performance inference uses the first 13 completed folds of a pre-generated frozen participant order, not all 38 usable recordings. No test participant was added, removed, replaced, or selected based on held-out performance, and the scientific configuration remained fixed during evidence accumulation. The all-38-recording pass verifies preprocessing applicability only and is not treated as 38-fold model evidence. External replication and broader participant-level testing remain desirable.

Second, the official Nightbeat comparison uses the official algorithm under the unified 60-s/15-s protocol rather than reproducing every detail behind the published 0.88 bpm result. Third, 93.68% of SeqSmoother endpoints align to an official Nightbeat candidate timestamp, and Nightbeat retains 77.76% of the aligned candidates, corresponding to 72.85% final availability on the SeqSmoother-eligible out-of-fold grid. Strict paired analysis excludes non-aligned endpoints, so absolute error must be interpreted jointly with prediction availability.

Fourth, NightbeatDB is dominated by typical sleeping HR in a limited physiological range. The present study does not establish performance for arrhythmia, markedly elevated nocturnal HR, cardiovascular disease cohorts, or other clinical sleep populations not adequately represented in the dataset. Fifth, the sub-harmonic result is a pooled mechanistic diagnostic across overlapping windows from seven positive-event participants; AUROC 0.972 is not participant-level inference and does not assign a unique amount of total regression improvement to $\eta_{1/2}$.

Finally, the current evaluation is retrospective and does not establish prospective real-time deployment performance. The model estimates window-level HR rather than beat-to-beat RR/NN intervals; conventional HRV, sleep-stage classification, disease prediction, clinical decision-making, and population-cohort outcome analysis remain outside the validated scope. Large accelerometer-only cohorts are therefore future application domains rather than demonstrated deployment settings.

## VIII. Conclusion

Across the first 13 completed folds of a frozen participant-disjoint order, SeqSmoother achieved participant-macro MAE 1.601 bpm on all out-of-fold windows. On Nightbeat-retained matched intervals, Nightbeat achieved lower participant-macro MAE (0.615 versus 1.091 bpm), with a paired difference of +0.475 bpm (95% CI +0.315 to +0.638, $p = 0.00049$). Nightbeat retained 77.76% of aligned candidates, corresponding to 72.85% final availability on the SeqSmoother-eligible out-of-fold grid. Matched within-participant-centered correlations were 0.878 for SeqSmoother and 0.854 for Nightbeat and are interpreted descriptively. Separately, the proposed sub-harmonic ratio achieved pooled diagnostic AUROC 0.972 across 337 reference-defined candidate lock-on windows from seven participants. The evidence therefore supports complementary behavior in absolute accuracy, prediction availability, and temporal tracking rather than universal superiority of either approach. These results motivate future hybrid accelerometer-only designs that combine quality-gated signal processing with learned temporal estimation.

## References

[1] A. Doherty, D. Jackson, N. Hammerla, T. Plo¨tz, P. Olivier, M. Granat, T. White, V. van Hees, M. Trenell, C. Owen, S. Preece, R. Gillions, S. Sheard, T. Peakman, S. Brage, and N. Wareham, “Large scale population assessment of physical activity using wrist worn accelerometers: The uk biobank study,” *PLOS ONE*, vol. 12, 02 2017.

[2] O. Inan, M. baran pouyan, A. Javaid, S. Dowling, M. Etemadi, A. Dorier, J. Heller, A. Bicen, S. Roy, and T. Marco, “Novel wearable seismocardiography and machine learning algorithms can assess clinical status of heart failure patients,” *Circulation: Heart Failure*, vol. 11, 01 2018.

[3] M. Conti, C. Aironi, S. Orcioni, R. Seepold, M. Gaiduk, and N. Mart´ınez Madrid, “Heart rate detection with accelerometric sensors under the mattress,” vol. 2020, 07 2020, pp. 4063–4066.

[4] R. G. Weaver, M. V. S. Chandrashekhar, B. Armstrong, J. W. White III, O. Finnegan, A. B. Cepni, S. Burkart, M. Beets, E. L. Adams, M. de Zambotti, G. J. Welk, S. Nelakuditi, D. Brown III, R. Pate, Y. Wang, R. Ghosal, Z. Zhong, and H. Yang, “Jerks are useful: extracting pulse rate from wrist-placed accelerometry jerk during sleep in children,” *Sleep*, vol. 48, no. 2, p. zsae099, 02 2025.

[5] J. Zschocke, J. Leube, M. Glos, O. Semyachkina-Glushkovskaya, T. Penzel, R. P. Bartsch, and J. W. Kantelhardt, “Reconstruction of pulse wave and respiration from wrist accelerometer during sleep,” *IEEE Transactions on Biomedical Engineering*, vol. 69, no. 2, pp. 830–839, 2022.

[6] J. Hernandez, D. J. McDuff, and R. W. Picard, “Bioinsights: Extracting personal data from “still” wearable motion sensors,” in *2015 IEEE BSN*.

[7] M. Moebus, L. Hauptmann, N. Kopp, B. Demirel, B. Braun, and C. Holz, “Nightbeat: Heart rate estimation from a wrist-worn accelerometer during sleep,” *IEEE Journal of Biomedical and Health Informatics*, vol. 29, pp. 870–877, 12 2024.

[8] V. Bieri, P. Streli, B. U. Demirel, and C. Holz, “BeliefPPG: Uncertainty-aware heart rate estimation from PPG signals via belief propagation,” in *Proceedings of the Thirty-Ninth Conference on Uncertainty in Artificial Intelligence*, ser. Proceedings of Machine Learning Research, R. J. Evans and I. Shpitser, Eds., vol. 216. PMLR, 31 Jul–04 Aug 2023, pp. 173–183. [Online]. Available: https://proceedings.mlr.press/v216/bieri23a.html

[9] S. Hochreiter and J. Schmidhuber, "Long short-term memory," *Neural Computation*, vol. 9, pp. 1735–1780, 11 1997.

[10] K. He, X. Zhang, S. Ren, and J. Sun, "Deep residual learning for image recognition," in *2016 IEEE Conference on Computer Vision and Pattern Recognition (CVPR)*, 2016, pp. 770–778.

[11] A. Vaswani, N. Shazeer, N. Parmar, J. Uszkoreit, L. Jones, A. Gomez, L. Kaiser, and I. Polosukhin, "Attention is all you need," 06 2017.

[12] M. Zhang, L. Qiu, Y. Chen, S. Yang, Z. Zhang, and L. Wang, "A conv-transformer network for heart rate estimation using ballistocardiographic signals," *Biomedical Signal Processing and Control*, vol. 80, p. 104302, 2023.

[13] A. K. Gupta, R. Kumar, L. Birla, and P. Gupta, "Radiant: Better rppg estimation using signal embeddings and transformer," in *Proceedings of the IEEE/CVF Winter Conference on Applications of Computer Vision (WACV)*, 2023, pp. 4976–4986.

[14] C. Kechris, J. Dan, J. Miranda, and D. Atienza, "Kid-ppg: Knowledge informed deep learning for extracting heart rate from a smartwatch," *IEEE Transactions on Biomedical Engineering*, vol. 72, no. 3, pp. 870–877, 2025.

[15] M. Risso, A. Burrello, D. J. Pagliari, S. Benatti, E. Macii, L. Benini, and M. Pontino, "Robust and energy-efficient ppg-based heart-rate monitoring," in *2021 IEEE International Symposium on Circuits and Systems (ISCAS)*, 2021, pp. 1–5.